\documentclass[letterpaper, 10 pt, conference]{ieeeconf}

\IEEEoverridecommandlockouts

\usepackage{cite}
\usepackage{xcolor}
\usepackage{acronym}
\usepackage{caption}
\usepackage{listings}
\usepackage{graphicx}
\usepackage{textcomp}
\usepackage{multirow}
\usepackage{booktabs}
\usepackage{subcaption}
\usepackage{algorithmic}
\usepackage{amsmath,amssymb,amsfonts}

\acrodef{VSLAM}{Visual SLAM}
\acrodef{FoV}{Field of View}
\acrodef{CV}{Computer Vision}
\acrodef{STD}{Standard Deviation}
\acrodef{LSD}{Line Segment Detector}
\acrodef{ROS}{Robot Operating System}
\acrodef{RGB-D}{Red Green Blue-Depth}
\acrodef{ATE}{Absolute Trajectory Error}
\acrodef{RMSE}{Root Mean Square Deviation}
\acrodef{CNN}{Convolutional Neural Network}
\acrodef{LiDAR}{Light Detection And Ranging}
\acrodef{ORB}{Oriented FAST and Rotated BRIEF}
\acrodef{API}{Application Programming Interface}
\acrodef{SLAM}{Simultaneous Localization and Mapping}

\newcommand{\wrt}{w.r.t. }
\newcommand{\orb}{ORB-SLAM 2.0 }

\newcommand{\ie}{\textit{i.e., }}
\newcommand{\ssslam}{{S$^{3}$LAM}}
\newcommand{\etal}{\textit{et al. }}

\def\BibTeX{{\rm B\kern-.05em{\sc i\kern-.025em b}\kern-.08em
    T\kern-.1667em\lower.7ex\hbox{E}\kern-.125emX}}
\begin{document}

\title{\LARGE \bf Marker-based Visual SLAM leveraging Hierarchical Representations}


\author{
    Ali Tourani$^{1}$, Hriday Bavle$^{1}$, Jose Luis Sanchez-Lopez$^{1}$, Rafael Muñoz Salinas$^{2}$, and Holger Voos$^{1}$ 
    \thanks{$^{1}$Authors are with the Automation and Robotics Research Group, Interdisciplinary Centre for Security, Reliability, and Trust (SnT), University of Luxembourg, Luxembourg. Holger Voos is also associated with the Faculty of Science, Technology, and Medicine, University of Luxembourg, Luxembourg. \tt{\small{\{ali.tourani, hriday.bavle, joseluis.sanchezlopez, holger.voos\}}@uni.lu}}
    \thanks{$^{2}$Author is with the Department of Computer Science and Numerical Analysis, Rabanales Campus, University of Córdoba, Spain. {\tt\small rmsalinas@uco.es}}
    \thanks{This work was funded by the Institute of Advanced Studies (IAS) of the University of Luxembourg (project TRANSCEND), the European Commission Horizon2020 research and innovation program under the grant agreement No 101017258 (SESAME), and the Luxembourg National Research Fund (FNR) 5G-SKY project (ref. C19/IS/13713801).}
    \thanks{For the purpose of Open Access, the author has applied a CC BY public copyright license to any Author Accepted Manuscript version arising from this submission.}
    \thanks{The authors would like to thank Deniz Işınsu Avşar and Jan P.F. Lagerwall, both from the Department of Physics \& Materials Science (DPHYMS) of the University of Luxembourg, for their efforts and valuable remarks on employing invisible fiducial markers as a futuristic scenario.}
}

\maketitle

\begin{abstract}
Fiducial markers can encode rich information about the environment and can aid \ac{VSLAM} approaches in reconstructing maps with practical semantic information.
Current marker-based \ac{VSLAM} approaches mainly utilize markers for improving feature detections in low-feature environments and/or for incorporating loop closure constraints, generating only low-level geometric maps of the environment prone to inaccuracies in complex environments.     
To bridge this gap, this paper presents a \ac{VSLAM} approach utilizing a monocular camera along with fiducial markers to generate hierarchical representations of the environment while improving the camera pose estimate. 
The proposed approach detects semantic entities from the surroundings, including walls, corridors, and rooms encoded within markers, and appropriately adds topological constraints among them.
Experimental results on a real-world dataset collected with a robot demonstrate that the proposed approach outperforms a traditional marker-based \ac{VSLAM} baseline in terms of accuracy, given the addition of new constraints while creating enhanced map representations.
Furthermore, it shows satisfactory results when comparing the reconstructed map quality to the one reconstructed using a \acs{LiDAR} SLAM approach.
\end{abstract}



\section{Introduction}
\label{sec_intro}


\acf{VSLAM} systems can employ a wide range of vision sensors, such as monocular, stereo, \ac{RGB-D}, omnidirectional, and event-based cameras to estimate the environmental map while localizing the camera \cite{cadena2016past}. The primary advantage of vision sensors is that they need low-cost hardware to supply rich visual and semantic information from surroundings for various tasks \cite{filipenko2018comparison}. 
Semantic data, which refers to high-level information acquired from the environment, make \ac{VSLAM} tasks more robust and expand the range of applications that can employ the reconstructed maps \cite{cadena2016past}, \cite{han2020dynamic}.
For instance, robots may need to identify objects in the scene for path planning and dynamic object removal, which can be resolved using semantic data \cite{wen2021semantic}.
In this regard, utilizing fiducial markers is one of the possible approaches to encoding semantic information into the environment due to their rich and better-defined features \cite{zhao2019visual}.
In addition, they can assist \ac{VSLAM} frameworks by providing accurate pose estimation due to their unique patterns, supplying reliable features in low-texture environments, and enabling loop closure detection based on marker identifiers.
While most fiducial markers are visible to humans and, thus, visually polluting the environment, we have recently proposed novel \textit{invisible fiducial markers} \cite{agha2022unclonable} that have the potential to be seamlessly integrated into the environment. Nevertheless, for the sake of simplicity, in this work, we use visible fiducial markers, leaving the use of \textit{invisible fiducial markers} as future works.
Recent works such as \cite{munoz2020ucoslam} and \cite{tagSLAM} propose \ac{VSLAM} approaches using fiducial markers but do not encode them with meaningful semantic information, creating purely geometric map representations leading to inaccuracies in camera pose estimates and the generated environmental map in the presence of complex environments.

\begin{figure}[t]
    \centering
    \includegraphics[width=1.0\columnwidth]{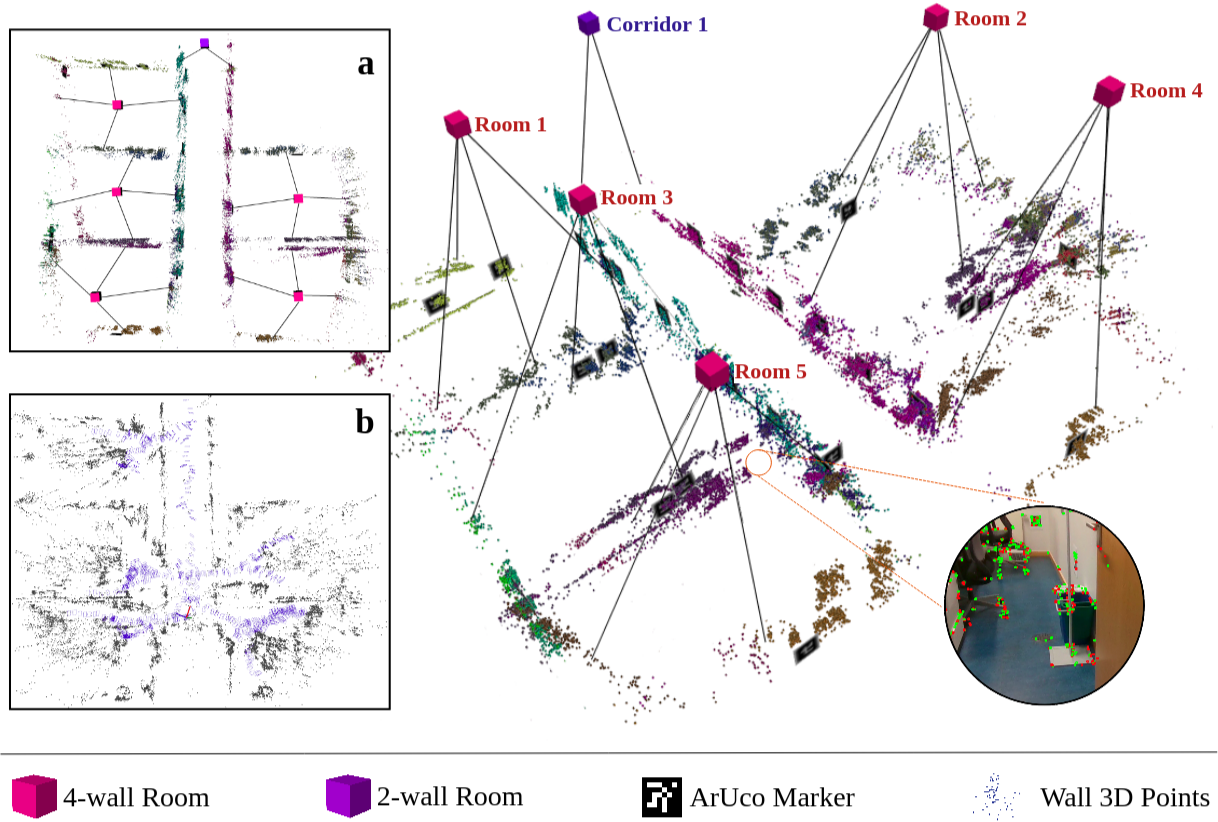}
    \caption{The final reconstructed map of the environment using the proposed method in a hierarchical representation. Accordingly, the primary map entities are detected 3D points extracted from the environment and visited ArUco markers. Various colors of the 3D points refer to distinct walls, and the line connecting them to the cubes shows their belonging to rooms, including corridors or four-wall rooms: a) the top view of the reconstructed map represented in 2D, b) keypoints and robot trajectory records that lead to map reconstruction.}
    \label{fig_overall}
\end{figure}

To fully leverage the potential of fiducial markers in accurately identifying both the semantic elements and their topological relationships while generating accurate and meaningful maps, this paper proposes a \ac{VSLAM} framework for monocular cameras that utilizes the data encoded in fiducial markers for enhanced map reconstruction, adding abstract semantic elements to the final map along with their topological relationships.
The system is built upon UcoSLAM \cite{munoz2020ucoslam}, which employs ArUco \cite{garrido2014automatic} markers and visual keypoints extracted from natural landmarks reconstructing the geometric map of the environment using keypoints and markers.
Additionally, inspired by \textit{S-Graphs} \cite{s_graphs, bavle2022s}, leveraging high-level hierarchical representations for localization and mapping, the proposed approach adds extraordinary information in the form of walls, corridors, and rooms considering the data encoded in the ArUco markers attached to the environment.
A sample map reconstructed by the proposed approach and its hierarchical representation is demonstrated in Fig.~\ref{fig_overall}. 

Herewith, the main contributions of the paper are summarized below:
\begin{itemize}
    \item An extension of marker-based \ac{VSLAM} to reconstruct environmental maps with high-level semantic features, including walls, corridors, and rooms, encoded within the markers,
    \item The design of novel geometric constraints, namely \textit{marker-to-wall} and \textit{wall-to-room}, to improve the map quality and reduce camera localization errors,    
    \item Validation of the proposed method using a real-world indoor dataset showing improved performance over marker-based VSLAM baseline.
\end{itemize}

The rest of the paper is organized as follows:
Section~\ref{sec_related} explores previous efforts in the \ac{VSLAM} domain, targeting the usage of semantic data for improved reconstructed maps.
Section~\ref{sec_proposed} introduces the proposed approach and its various modules in detail.
In Section~\ref{sec_evaluation}, the evaluation criteria and discussions over the effectiveness of the proposed method are provided.
The paper finally concludes in Section~\ref{sec_conclusions}.
\section{Related Works}
\label{sec_related}

\ac{VSLAM} approaches have matured over the past years and many researchers contributed to this community by proposing efficient solutions to unresolved challenges in this area.
The authors of this paper also presented a comprehensive survey on diverse \ac{VSLAM} state-of-the-art works and studied their trade-offs in \cite{tourani2022visual}.
Accordingly, many recent approaches focus on improving the \textit{front-end} thread of the \ac{VSLAM} systems for better pose estimation and visual odometry \cite{kazerouni2022survey}.
Authors in \cite{munoz2020ucoslam} proposed UcoSLAM that employs both natural and artificial landmarks in the environment to reconstruct the map.
It works in keypoints-only, markers-only, and mixed modes and improves the global map builder and loop closure detector modules in case ArUco markers are found in the camera's field of view.
Another marker-based approach titled TagSLAM \cite{tagSLAM} was proposed by Pfrommer and Daniilidis, which uses the capability of AprilTag \cite{olson2011apriltag} fiducial markers for \acs{SLAM} tasks.
Although the proposed approaches perform well in many scenarios, they cannot obtain any semantic information from the environment for an improved map representation.
Gomez-Ojeda \etal \cite{gomez2019pl} introduced PL-SLAM, which uses \ac{ORB} and \ac{LSD} algorithms to extract visual features of points and lines in stereo vision cameras.
However, this approach is computationally intensive in feature tracking and extracting data from structural lines.
Bruno and Colombini \cite{bruno2021lift} proposed LIFT-SLAM, which integrates deep learning-based and conventional geometry-based feature descriptors.
Despite the robust feature detection and tracking performance, the method uses an un-optimized \ac{CNN}, which leads to almost real-time execution.

Utilizing semantic data is another topic that attracted great attention in recent years.
Gonzalez \etal \cite{gonzalez2021s3lam} presented \ssslam, a monocular approach based on \orb \cite{mur2017orb} that uses semantic segmentation of generic objects and structures.
Their approach fits a plane using the triangulated 3D points for each a priori planar cluster and uses a \ac{CNN}-based panoptic segmentation framework titled \textit{Detectron2} \cite{wu2019detectron2}.
Likewise, YOLO-SLAM \cite{wu2022yolo} is another \ac{CNN}-based approach that couples geometric constraints with the Darknet19-YOLOv3 object detector to generate semantic information.
However, these solutions may suffer from performance degradation in recognizing small or non-regular objects and planes.
QuadricSLAM \cite{nicholson2018quadricslam} is another approach that models objects found in the environment by employing quadric representations as object descriptions and estimates quadric parameters using object detection bounding boxes.
This monocular \ac{VSLAM} represents an object's size, position, and orientation for map reconstruction.
The main challenge of QuadricSLAM is the initialization stage, where a variety of viewing angles of the object is required.
Hosseinzadeh \etal \cite{hosseinzadeh2019real} proposed another monocular \ac{VSLAM} system that inserts visited objects detected by a deep learning-based object detector and the dominant structure of the scene using a planar landmark detector to the map.
Despite the robust performance, estimating the prior shape of the objects leads to high computational complexity.
Another semantic \ac{VSLAM} approach titled Blitz-SLAM \cite{fan2022blitz} combines the original masks and depth information of the objects and eliminates noise blocks formed by dynamic objects.
Authors in \cite{sun2022solo} proposed a parallel framework titled SOLO-SLAM, which uses semantic attributes of map points and geometric constraints to filter dynamic objects and reconstruct an improved map.
The main challenge here is losing a great deal of information in low-dynamic scenarios.
MISD-SLAM \cite{you2022misd} is another semantic approach equipped with real-time instance segmentation based on a pre-trained \ac{CNN} and dynamic feature removal.
However, the algorithm removes objects that are static only in some frames.

In contrast with the mentioned works, the proposed approach creates a map of the environment by utilizing semantic data encoded within makers instead of employing any object detector.
Accordingly, walls and room constraints are estimated based on mathematical modeling obtained from keypoints and markers data, which does not significantly affect the performance of the system.
Additionally, the accuracy of the final reconstructed map of the environment is comparable with the one from \ac{LiDAR}-based approaches, demonstrating the proposed method's potential.
\section{Proposed Method}
\label{sec_proposed}

\begin{figure}[t]
    \centering
    \includegraphics[width=1.0\columnwidth]{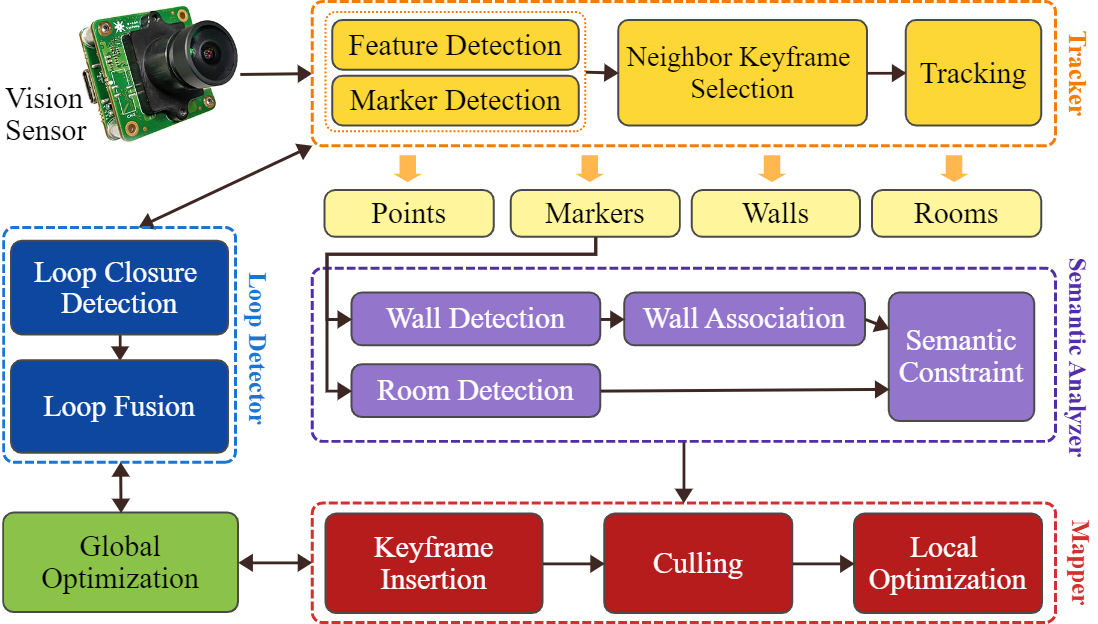}
    \caption{The overall pipeline of the proposed approach. The main contributions of the proposed method are adding the \textit{Semantic Analyzer} module and modifying the output of the \textit{Tracker} to provide proper feed to it. Moreover, the \textit{Global/Local Optimizer} modules have also been modified for better map reconstruction.}
    \label{fig_pipeline}
\end{figure}

The pipeline of the proposed method is depicted in Fig.~\ref{fig_pipeline}.
Accordingly, the frames captured by a monocular camera are passed to the system for processing.
The \textit{tracker} module detects \ac{ORB} visual features and ArUco markers as the primary inputs of the system.
Keyframes are selected and tracked over time, and the output contains 3D points, detected markers, walls, and rooms visited in the scene.
The \textit{semantic analyzer} module performs wall and room detection tasks using the pose of the markers detected in the \textit{tracking} stage.
Additionally, a \textit{mapper} module handles adding new vertices and edges to the map and improves the final map for an optimized reconstruction.
Finally, a marker-based \textit{loop detector} module checks whether the markers in the current frame have been seen before and triggers the \textit{global optimizer} to enhance the reconstructed map.

In order to develop a \ac{VSLAM} framework with richer reconstructed maps, the proposed approach utilizes UcoSLAM as the baseline and modifies its components to be empowered with a semantic data analysis procedure.
The mentioned modifications enable the detection of \textit{walls} and different types of \textit{rooms} as two semantic concepts in the environment using ArUco markers.
Furthermore, as these fiducial markers are used for camera pose estimation, they can store additional encoded semantic data, which makes them an ideal alternative according to their unique identification properties.
This method also aspires to employ visual data to represent the environment and the robot's pose in a single optimizable graph with an approach comparable to \textit{S-Graphs} \cite{bavle2022situational}.
Thus, instead of odometry readings and extracting planar surfaces using \ac{LiDAR} scans introduced by \textit{S-Graphs} and \textit{S-Graphs+}~\cite{bavle2022s}, the proposed approach calculates the location of walls and room variations using the fiducial markers attached to them.

\subsection{Overview}
\label{sec_proposed_overview}

The pipeline of the proposed method at time \(t\) can be referred to four main coordinate systems: the odometry frame of reference $O$, the camera coordinate system $C_t$, the marker coordinate system $M_t$, and the global coordinate system $G_t$.
As the primary sensor of the system, a monocular camera acquires a set \(\mathbf{F}=\{\mathbf{f}\}\) of frames \(\mathbf{f}=\displaystyle\{t,\mathbf{T},\mathbf{\delta}\}\), where \(\mathbf{T} \in SE(3)\) is the camera's pose obtained from the transformation of $C_t$ to $G_t$, and \(\delta\) is the set of camera intrinsic parameters.
By processing each camera frame using \ac{ORB} keypoint detector and feature extractor, a set of keypoints \( \mathbf{g} =\{l, u, \mathbf{d}\} \) are extracted in which \(l\) is the subsampling level of the image, \(u\) is the pixel coordinates for upsampling \wrt the first level, and \(\mathbf{d}=(d_1...d_n)|d_i\in[0,1]\) is the descriptor vector with length $n$.
Accordingly, the final constructed map of the environment \(\mathbf{E}\) will be represented as:

\begin{equation}
    \mathbf{E}=\{\mathbf{K}, \mathbf{P}, \mathbf{M}, \mathbf{W}, \mathbf{R}\}
\end{equation}

\noindent where \(\mathbf{K}=\{k\}\subset\mathbf{F}\) is the set of keyframes and \(\mathbf{P}=\{\mathbf{p}\}\) represents the set of feature points \(\mathbf{p}=\{\mathbf{x},\mathbf{v},\mathbf{\hat{d}}\}\) extracted from the environment with their corresponding 3D positions \(\mathbf{x}\in\mathbb{R}^3\), viewing direction \(\mathbf{v}\in\mathbb{R}^3\), and descriptor \(\mathbf{\hat{d}}\).
Additionally, \(\mathbf{M}=\{\mathbf{m}\}\) is the set of ArUco markers detected in the environment, in which each marker \(\mathbf{m}=\{s,\mathbf{p},\mathbf{cc}\}\) holds marker size (\ie length) \(s\in\mathbb{R}\), marker pose \(\mathbf{p} \in SE(3)\) calculated from $M_t$ to $G_t$, and corner coordinates \(\mathbf{cc}=(c_1...c_4)|c_i\in\mathbb{R}^3\) values.
The set of walls detected from the environment is represented by \(\mathbf{W}=\{\mathbf{w}\}\), in which each wall \(\mathbf{w}=\{\mathbf{q},\mathbf{m_w}\}\) holds the wall equation \(\mathbf{q}\in\mathbb{R}^4\) and the attached markers list \(\mathbf{m_w}\subset\mathbf{M}=(m_1...m_n)|m_i\in\mathbb{N} \) where $m_i$ represents ArUco \textit{marker-id}.
Similarly, \(\mathbf{R}=\{\mathbf{r}\}\) refers to the set of rooms found in the environment, where each room \(\mathbf{r}=\{\mathbf{r_c},\mathbf{r_w}\}\) contains the room center point \(\mathbf{r_c}\in\mathbb{R}^3\) and the wall list \(\mathbf{r_w}\subset\mathbf{W}=(w_1...w_n)|w_i\in\mathbb{N} \) that comprise the room.
\subsection{Semantic Entities}
\label{sec_proposed_entities}

The proposed approach adds two semantic entities to construct a richer map, including \textit{walls} and \textit{rooms}.
As detecting the mentioned entities is accomplished differently, the procedures are discussed in detail in this section.

\textbf{Walls.}
The procedure to find the walls as plane-shaped entities on which the markers and their surrounding features are located is predominantly done using the pose information provided by detected ArUco markers.
This approach assumes all fiducial markers are placed directly on the walls, not other elements such as radiators or drawer units.
Thus, the equations of the walls are obtained based on the attached markers’ poses.   
The process of detecting walls in the scene occurs whenever a marker is being visited in a recent keyframe, making detecting and mapping the walls at a given time more efficient.

To add walls to the final map, each wall plane \(\boldsymbol{w}_i\) is extracted in the global coordinate system \(^{G} \boldsymbol{w}_{i} = \begin{bmatrix} ^{G}\boldsymbol{n}_i & ^{G}d\end{bmatrix}\) with a normal vector \(^{G} \boldsymbol{n}_{i} = \begin{bmatrix} n_{x} & n_{y} & n_{z} \end{bmatrix}^{T}\).
The vertex node of the wall is factored in the graph as ${^G}\boldsymbol{w_i} = [^G\phi, ^G\theta, {^G}d]$, where $^M\phi$ and $^M\theta$ refer to the azimuth and elevation of the wall in $G_t$.
For each marker $^{G} \boldsymbol{m}_i$ attached to the wall $^{G} \boldsymbol{w}_i$ cost function can be defined as:  

\begin{equation}
    c_{w_i}(^G\boldsymbol{w}_i, ^G\boldsymbol{m}_i) = \| [^{M}\delta\phi_{w_{i_{m_i}}}, ^{M}\delta\theta_{w_{i_{m_i}}}, ^{M}d_{w_i} ]^{T} \|^2_{\mathbf{\Lambda}_{\boldsymbol{\tilde{w}}_{i}}}
    \label{eq_wall_marker_edge}
\end{equation}

\noindent where $^{M} \delta\phi_{w_{i_{m_i}}}$ difference between the azimuth angle of the wall $w_i$ and its marker $m_i$ converted to its marker frame $M_i$, \(^{M} \delta\theta_{w_{i_{m_i}}}\) is the difference in the elevation angles, while $^{M} d_{w_i}$ being the perpendicular distance between the wall and the marker, which should be zero the given marker-wall pair. 
\newline
\newline
\textbf{Rooms.}
Another semantic entity that has been considered in this work is the room.
Since perceiving a room can be difficult due to various configurations and structures, the proposed approach employs the data encoded in ArUco markers attached to the room’s walls to detect the mentioned semantic entity.
In this regard, a dictionary containing the rooms in the environment and the fiducial markers attached to their walls are provided to feed the framework.
Note that the only information encoded in the dictionary is the \textit{marker-ids} corresponding to a room, and no additional pose information is required to be encoded.
The proposed approach presents a modified version of the room segmentation process introduced in \textit{S-Graphs+} \cite{bavle2022s}, where the markers play a vital role.
Hence, two room types titled \textit{"two-wall room"} and \textit{"four-wall room"} have been considered in this work:

\textbf{\textit{Two-wall Rooms (Corridors):}}
In this case, only two parallel walls of a room are labeled with fiducial markers.
This scenario is proper for detecting corridors or rooms with undetectable/unreachable walls in the scene.
Consequently, a room $^{G}\mathbf{r}_{x}=[{^G}{\mathbf{w}_{x_{a_1}}},{^G}{\mathbf{w}_{x_{b_1}}}]$ contains $x$-wall planes parallel to the $x$-axis while $^{G}\mathbf{r}_{y}=[{^G}{\mathbf{w}_{y_{a_1}}},{^G}{\mathbf{w}_{y_{b_1}}}]$ contains $y$-wall planes parallel planes to $y$-axis. 
To compute the center of a two-wall room $^{G}\mathbf{r}_{x_{i}}$, the two $x$-wall plane equations are utilized along with the center point ${^G}{\mathbf{c}_{i}}$ of the marker $\mathbf{m}_{i}$ as follows:

\begin{gather}
    \resizebox{1.\hsize}{!}{${^G}{\mathbf{k}_{x_i}} = \frac{1}{2} 
    \big[ \lvert {{^G}{d_{x_{a_1}}} \rvert} \cdot {^G}{\mathbf{n}_{x_{a_1}}} - {\lvert {^G}{d_{x_{b_1}}} \rvert} \cdot {^G}{\mathbf{n}_{x_{b_1}}} \big] + \lvert {^G}{{d_{x_{b_1}}} \rvert} \cdot {^G}{\mathbf{n}_{x_{b_1}}} \nonumber$} \\
    {^G}{\boldsymbol{\eta}_{x_i}} = {^G}{\hat{\mathbf{k}}_{x_i}} + \big[ {^G}{\mathbf{c}_i} - [\ {^G}{\mathbf{c}_i} \cdot {^G}{\hat{\mathbf{k}}_{x_i}} ] \ \cdot \hat{{^G}{\mathbf{k}}_{x_i}} \big]
    \label{eq_two_wall_room_center}
\end{gather}

\noindent where ${^G}{\boldsymbol{\eta}_{x_i}}$ is the center point of the two-wall room  $^{G}\mathbf{r}_{x_{i}}$ and ${^G}{\hat{\mathbf{k}}_{x_i}}$ is acquired from ${^G}{\hat{\mathbf{k}}_{x_i}} = {^G}{\mathbf{k}_{x_i}} / \| {^G}{\mathbf{k}_{x_i}}\|$.
The center point ${^G}{\mathbf{c}_{i}}$ of the marker is obtained using the marker pose in frame $G$.
It should be noted that a two-wall room center in the $y$ direction can be calculated analogously.


A two-wall room node is initialized using the room center, and the cost function to minimize the two-wall room node and their corresponding wall planes are defined below:

\begin{multline}
    \label{eq_two_wall_room_node}
    c_{\boldsymbol{r}_{{x_i}}}({^G}{\boldsymbol{r}_{{x_i}}},\big[{^G}{\mathbf{w}_{x_{a_1}}}, {^G}{\mathbf{w}_{x_{b_1}}}, {^G}{\textbf{c}_i}\big]) \\ = \sum_{t=1,i=1}^{T,K} \| {^G}{\hat{\boldsymbol{\eta}}_{x_i}} - f({^G}{\tilde{\mathbf{w}}_{x_{a_1}}}, {^G}{\tilde{\mathbf{w}}_{x_{b_1}}}, {^G}{\textbf{c}_i}) \| ^2_{\boldsymbol{\Lambda}_{\boldsymbol{\tilde{\boldsymbol{r}}}_{i,t}}}
\end{multline}

\noindent where $f({^G}{\tilde{\mathbf{w}}_{x_{a_1}}}, {^G}{\tilde{\mathbf{w}}_{x_{b_1}}}, {^G}{\textbf{c}_i})$ maps the wall planes to the center point of the room using Eq.~\ref{eq_two_wall_room_center}.

\textbf{\textit{Four-wall Rooms:}}
This scenario refers to rooms with four walls labeled with fiducial markers. 
In this regard, a four-wall room contains four wall planes as $^{G}\mathbf{r}_{i}=[{^G}{\mathbf{w}_{x_{a_1}}} {^G}{\mathbf{w}_{x_{b_1}}} {^G}{\mathbf{w}_{y_{a_1}}} {^G}{\mathbf{w}_{y_{b_1}}}]$ forming the room.
The center point of this variant of rooms can be computed using the equation below:

\begin{gather}
    \resizebox{1.\hsize}{!}{$
    {^G}{\mathbf{q}_{x_i}} = \frac{1}{2} 
    \big[ \lvert {{^G}{d_{x_{a_1}}} \rvert} \cdot {^G}{\mathbf{n}_{x_{a_1}}} - {\lvert {^G}{d_{x_{b_1}}} \rvert} \cdot {^G}{\mathbf{n}_{x_{b_1}}} \big] + \lvert {{^G}{d_{x_{b_1}}} \rvert} \cdot {^G}{\mathbf{n}_{x_{b_1}}} \nonumber$}
    \\
    \resizebox{1.\hsize}{!}{${^G}{\mathbf{q}_{y_i}} = \frac{1}{2} \big[ \lvert {{^G}{d_{y_{a_1}}} \rvert} \cdot {^G}{\mathbf{n}_{y_{a_1}}} - {\lvert {^G}{d_{y_{b_1}}} \rvert} \cdot {^G}{\mathbf{n}_{y_{b_1}}} \big] + \lvert {{^G}{d_{y_{b_1}}} \rvert} \cdot {^G}{\mathbf{n}_{y_{b_1}}} \nonumber$} \\
    {^G}{\boldsymbol{\rho}_i}  = {^G}{\boldsymbol{q}_{x_i}} + {^G}{\boldsymbol{q}_{y_i}}
    \label{eq_four_wall_room_center}
\end{gather}

\noindent where ${^G}{\boldsymbol{\rho}_{i}}$ is the center point of the four-wall room $^{G}\mathbf{r}_{i}$.
It should also be noted that Eq.~\ref{eq_four_wall_room_center} holds true when \mbox{$\lvert d_{x_1} \rvert > \lvert d_{x_2} \rvert$}.
The cost function to minimize four-wall room nodes and their corresponding wall plane set is similar to a two-wall room but with minor differences:

\begin{multline}
    \label{eq_four_wall_room_node}
    c_{\boldsymbol{\rho}} ({^G}{\boldsymbol{\rho}}, \big[ {^G}{\mathbf{w}_{x_{a_i}}}, {^G}{\mathbf{w}_{x_{b_i}}}, {^G}{\mathbf{w}_{y_{a_i}}}, {^G}{\mathbf{w}_{y_{b_i}}}\big]) \\ = \sum_{t=1, i=1}^{T, S} \| {^G}{\hat{\boldsymbol{\rho}}_i} - {{f({^G}{\tilde{\mathbf{w}}_{x_{a_i}}}, {^G}{\tilde{\mathbf{w}}_{x_{b_i}}}, {^G}{\tilde{\mathbf{w}}_{y_{a_i}}}, {^G}{\tilde{\mathbf{w}}_{y_{b_i}}})}} \| ^2_{\mathbf{\Lambda}_{\boldsymbol{\tilde{\boldsymbol{\rho}}}_{i,t}}}
\end{multline}

\noindent where $f({^G}{\tilde{\mathbf{w}}_{x_{a_i}}}, {^G}{\tilde{\mathbf{w}}_{x_{b_i}}}, {^G}{\tilde{\mathbf{w}}_{y_{a_i}}}, {^G}{\tilde{\mathbf{w}}_{y_{b_i}}})$ maps the four estimated wall planes to the center point of the four-wall room using Eq.~\ref{eq_four_wall_room_center}.
The above cost function maintains the structural consistency among the four walls forming the room.
\subsection{Final Graph}
\label{sec_proposed_graph}

\begin{figure}[t]
    \centering
    \includegraphics[width=1.0\columnwidth]{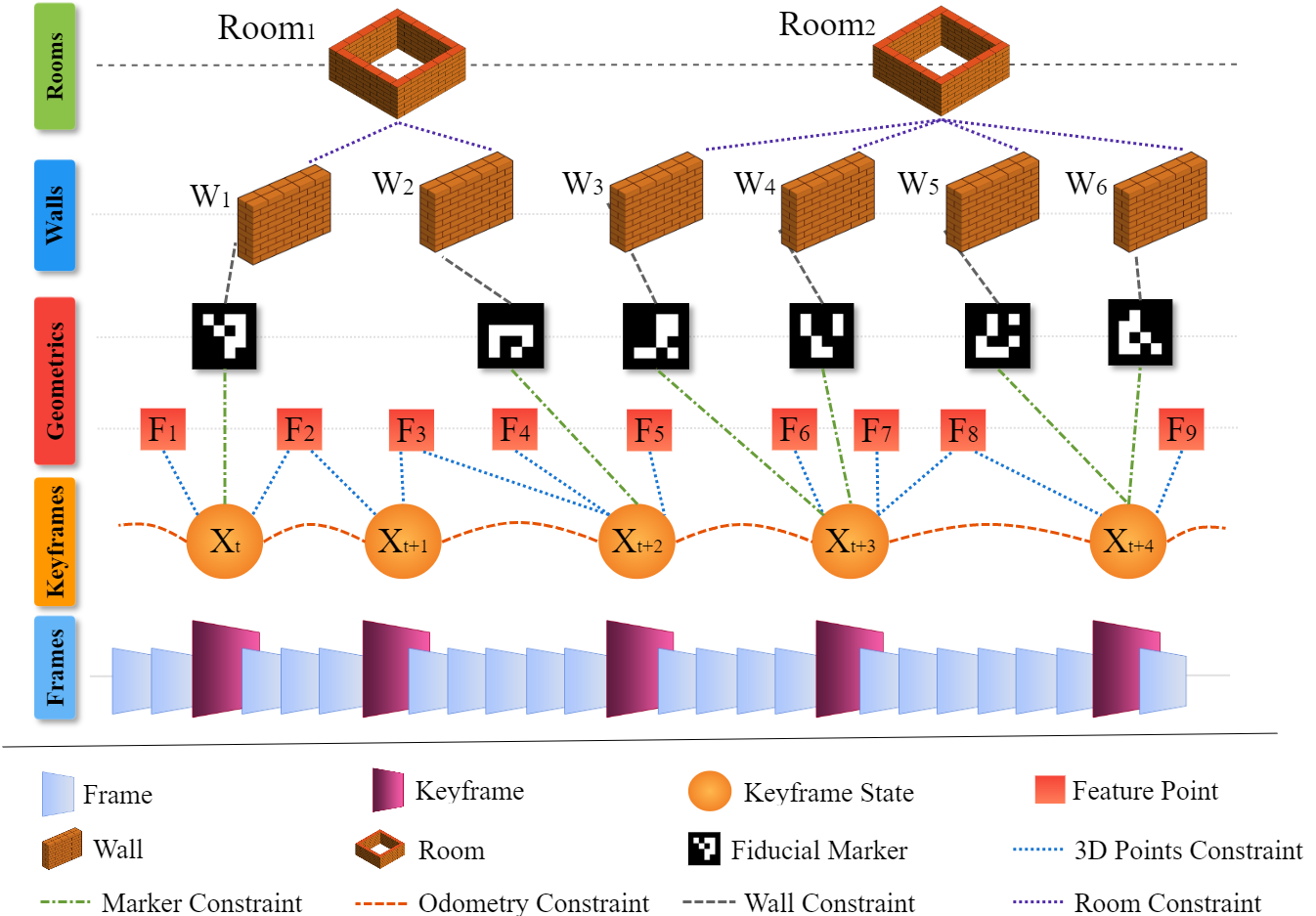}
    \caption{The graph representation of the hierarchical architecture of the proposed approach. In this regard, the map employs new semantic constraints, including walls and rooms, along with geometrical constraints for better map reconstruction.}
    \label{fig_graph}
\end{figure}

Fig.~\ref{fig_graph} depicts the structure of the final semantic graph produced by the proposed approach.
Accordingly, the keyframes extracted by the system are the primary sources of information that contain both visual feature points with their corresponding 3D coordinates and visited ArUco markers in the scene.
It should be noted that the constraint defined among the keyframes, feature points, and fiducial markers is utilized for computing the odometry and for detecting loop closure.
The wall-marker constraints are incorporated using Eq.~\ref{eq_wall_marker_edge}.
The topmost level of the graph retains rooms detected in the environment using the \textit{marker-ids} and the walls that hold those markers with constraints obtained following Eq.~\ref{eq_two_wall_room_node} for two-wall rooms and Eq.~\ref{eq_four_wall_room_node} for four-wall rooms. 

\section{Evaluation}
\label{sec_evaluation}

This section presents the proposed method's evaluation procedure in terms of accuracy and performance compared to the state-of-the-art approaches.
In this respect, various real-world scenario tests were performed using the proposed method, UcoSLAM \cite{munoz2020ucoslam} as the baseline methodology, and \textit{S-Graph+} \cite{bavle2022s} as a \ac{LiDAR}-based approach for providing ground truth measurements.
As \ac{LiDAR} sensor has less noise when compared with the monocular vision-based sensor, it can be safely considered as ground truth.

\subsection{Evaluation Setup}
\label{sec_eval_setup}

\begin{figure}[t]
    \centering
    \includegraphics[width=1.0\columnwidth]{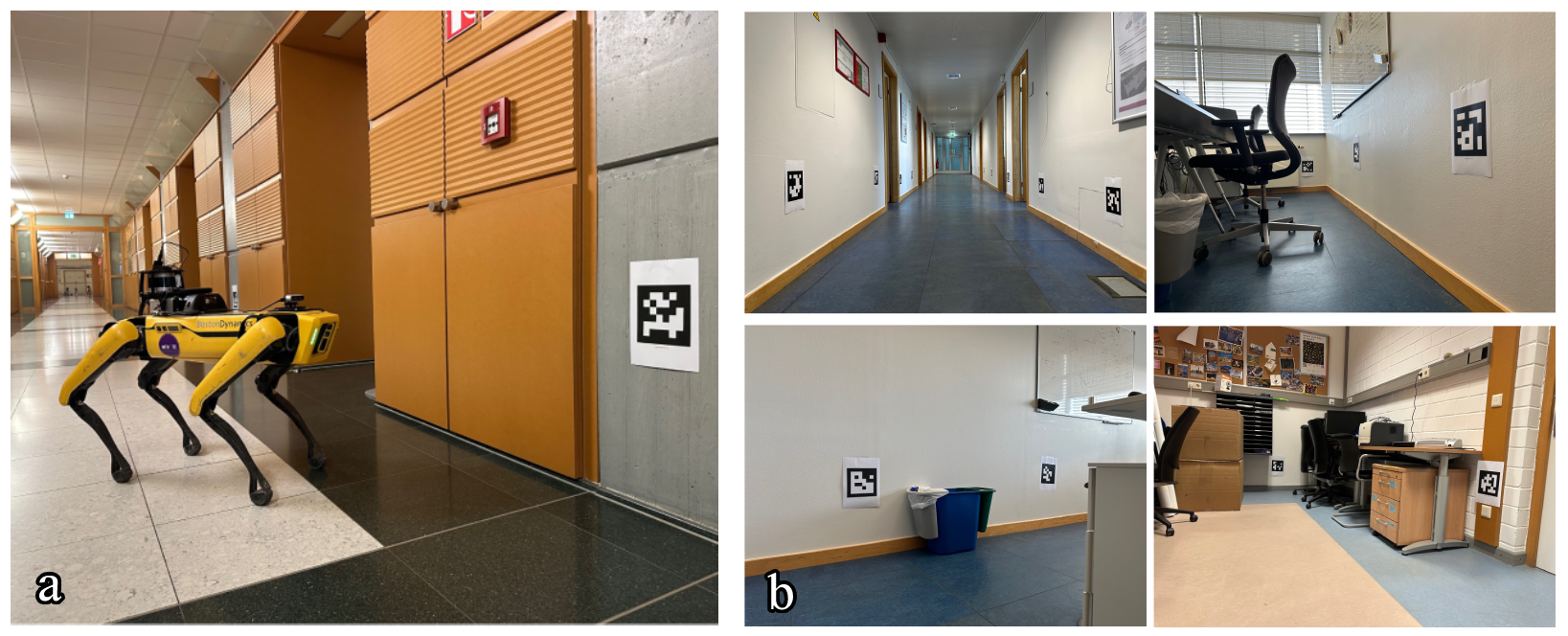}
    \caption{Collected dataset for evaluation of the method: a) the legged robot used for data collection, b) some instances of the dataset.}
    \label{fig_dataset}
\end{figure}

In order to evaluate the performance of the proposed approach in real-world circumstances, we mounted a \textit{Intel® RealSense™ Depth Camera D435} as the monocular sensor on a \textit{Boston Dynamics Spot®} robot and collected data from an indoor environment.
The robot functioned in different office zones of two different university buildings with various corridor and room setups, where the walls were labeled with printed \(17cm \times 17cm\) ArUco markers.
\textit{Marker-ids} of the ArUco markers placed in the environment, along with the room unique labels, were also stored in a database file and fed to the system.
Fig.~\ref{fig_dataset} demonstrates the robot employed for collecting data and some instances of the collected dataset.
Additionally, the characteristics of the dataset are presented in Table~\ref{tbl_dataset}.

\begin{table}[t]
    \scriptsize
    \centering
    \caption{The characteristics of the collected indoors dataset.}
    \begin{tabular}{l | c | c | p{3.2cm}}
        \toprule
        \textbf{\textit{Sequence$^{\mathrm{*}}$}} & \textbf{\textit{Duration}}& \textbf{\textit{\#Markers}}& \textbf{\textit{Description}} \\
        \midrule
            \textit{Seq-01} & 08m 53s & 35 & Two corridors connected via a landing without robot rotation \\
            \textit{Seq-02} & 09m 29s & 35 & Two corridors connected via a landing with robot rotation \\
            \textit{Seq-03} & 16m 05s & 26 & One corridor and five rooms \\
            \textit{Seq-04} & 10m 25s & 14 & One corridor and a two-doors room \\
            \textit{Seq-05} & 09m 37s & 20 & Four aisles connected to a main corridor \\
            \textit{Seq-06} & 23m 34s & 22 & Three corridors and three rooms \\
        \bottomrule
        \multicolumn{4}{l}{$^{\mathrm{*}}$data were stored as packages of \textit{rosbag} files.}
    \end{tabular}
    \label{tbl_dataset}
\end{table}

Evaluations have been conducted using a computer equipped with an \textit{11th Gen. Intel® Core™ i9 @2.60GHz} processor and 32 GigaBytes of memory.
Moreover, to better visualize the detected objects in the environment, including 3D points, ArUco markers, walls, and rooms, and provide a more accurate evaluation to compare the proposed method with other state-of-the-art approaches, a set of \ac{ROS} tools have been employed.
\subsection{Experimental Results}
\label{sec_eval_evaluation}

In order to demonstrate the accuracy of the proposed method compared to its baseline and ground truth, \ac{ATE} measurements have been employed in this paper.
Accordingly, the \ac{RMSE} and \ac{STD} values of the proposed and baseline approaches were compared to the ground truth, and the approach with less value is assumed to perform more accurately.

\begin{figure}[t]
     \centering
     \begin{subfigure}[t]{0.49\columnwidth}
         \centering
         \includegraphics[width=\columnwidth]{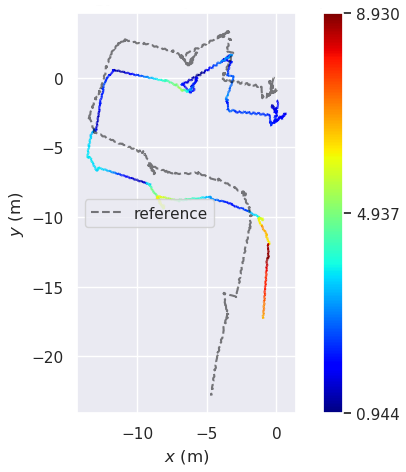}
         \caption{Proposed}
         \label{fig_eval_charts_proposed}
         \vspace{1em}
     \end{subfigure}
     \hfill
     \begin{subfigure}[t]{0.49\columnwidth}
         \centering
         \includegraphics[width=\columnwidth]{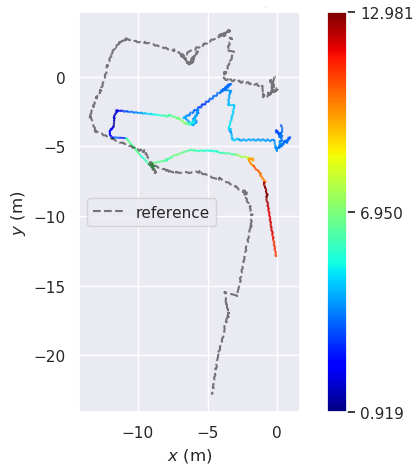}
         \caption{UcoSLAM \cite{munoz2020ucoslam}}
         \label{fig_eval_charts_uco}
     \end{subfigure}
     \caption{\acf{ATE} \wrt translation part in \textit{meters} on \textit{Seq-04} using different approaches.
     It can be seen that the room created by the proposed approach is more similar to the ground truth.}
     \label{fig_eval_charts}
\end{figure}

According to the evaluation results presented in Table~\ref{tbl_evaluation}, the proposed approach works better than its baseline in most of the cases.
The main reason for such improvement is the ability of the proposed method to add new constraints to the map and employ the association of semantic entities to enhance the reconstruction of the final map.
The mentioned improvement can also be seen in Fig.~\ref{fig_eval_charts}. 
Moreover, the proposed approach utilizes data association results and performs a local \textit{semantic loop closure} in addition to the marker-based loop closure detection when visiting a previously observed wall.
In this regard, if the equation of the currently visible wall was previously measured, \ie the wall was previously detected, a global optimization runs again to improve the reconstructed map.

\begin{figure*}[t]
     \centering
     \begin{subfigure}[t]{0.23\textwidth}
         \centering
         \includegraphics[width=\textwidth]{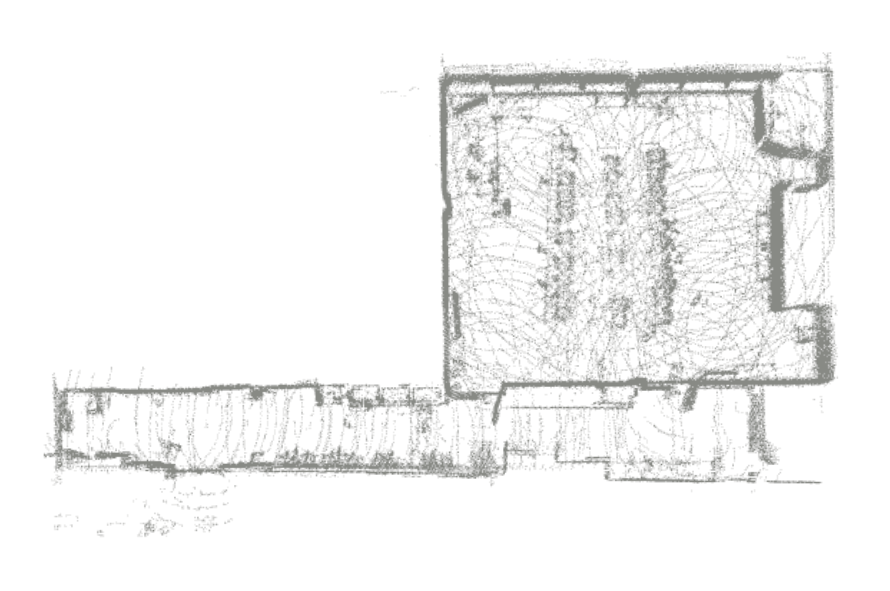}
         \caption{\textit{S-Graphs+} on \textit{Seq-04}}
         \label{fig_eval_sgraphs_s4}
     \end{subfigure}
     \hfill
     \begin{subfigure}[t]{0.23\textwidth}
         \centering
         \includegraphics[width=\textwidth]{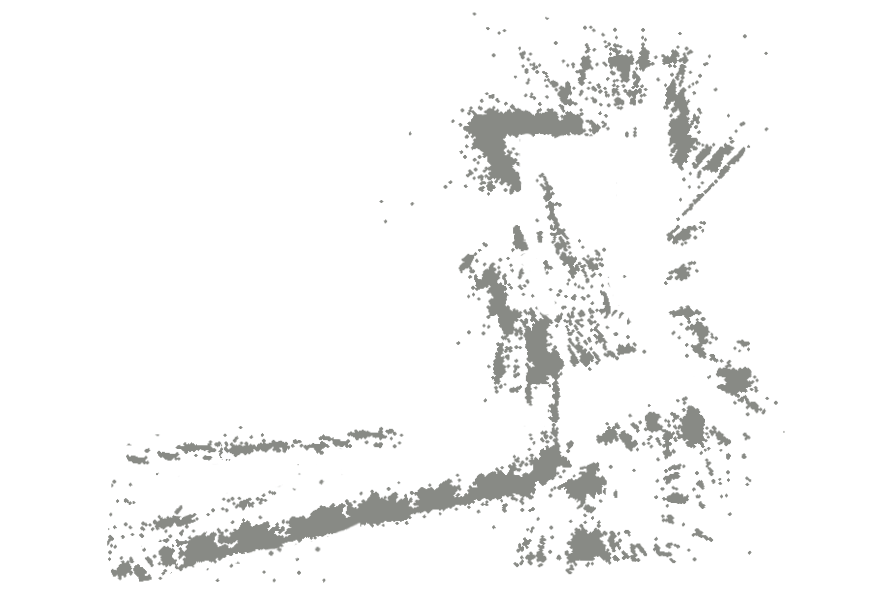}
         \caption{UcoSLAM \cite{munoz2020ucoslam} on \textit{Seq-04}}
         \label{fig_eval_uco_s4}
     \end{subfigure}
     \hfill
     \begin{subfigure}[t]{0.23\textwidth}
         \centering
         \includegraphics[width=\textwidth]{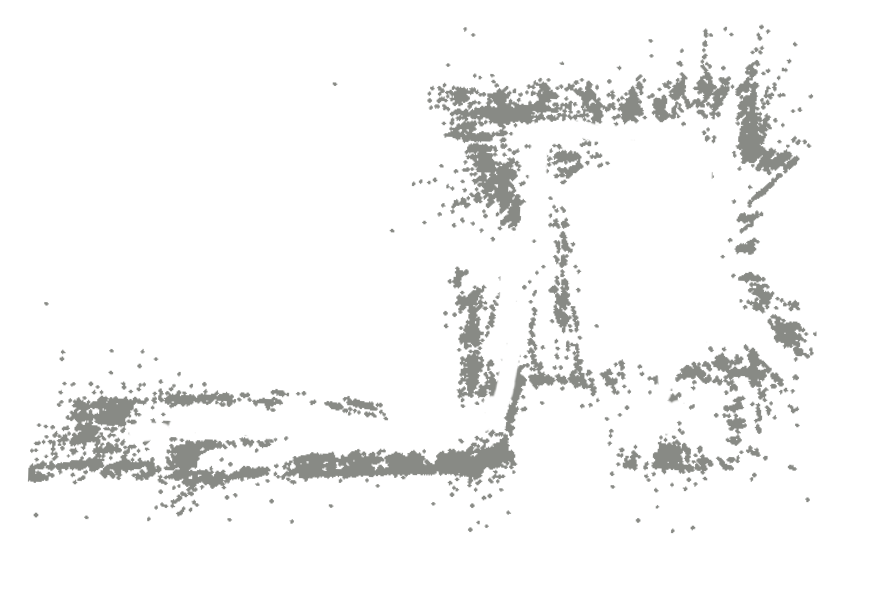}
         \caption{Proposed method on \textit{Seq-04}}
         \label{fig_eval_proposed_s4}
     \end{subfigure}
     \hfill
     \hfill
     \begin{subfigure}[t]{0.23\textwidth}
         \centering
         \includegraphics[width=\textwidth]{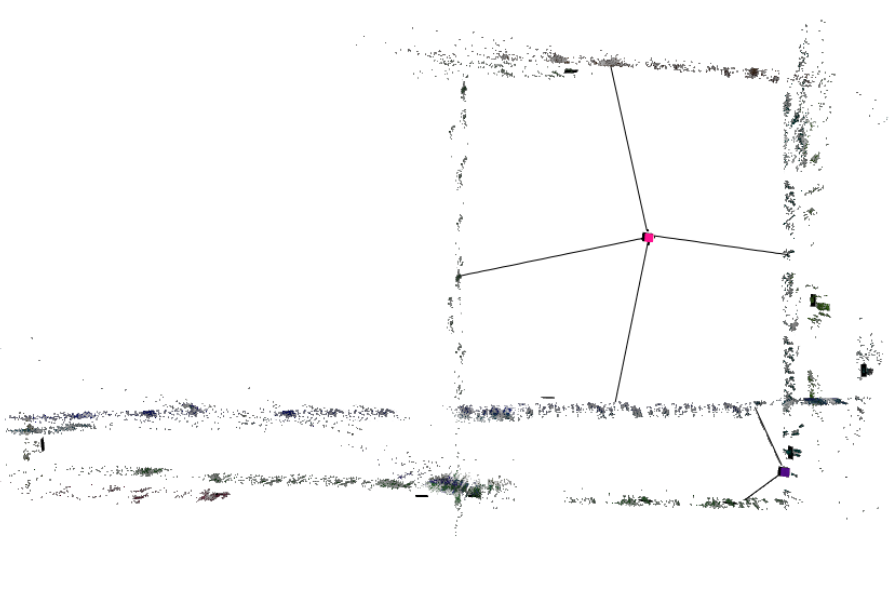}
         \caption{Proposed method on \textit{Seq-04} showing only wall 3D points}
         \label{fig_eval_semantic_s4}
         \vspace{0.5em}
     \end{subfigure}     
     \bigskip
     \begin{subfigure}[t]{0.23\textwidth}
         \centering
         \includegraphics[width=\textwidth]{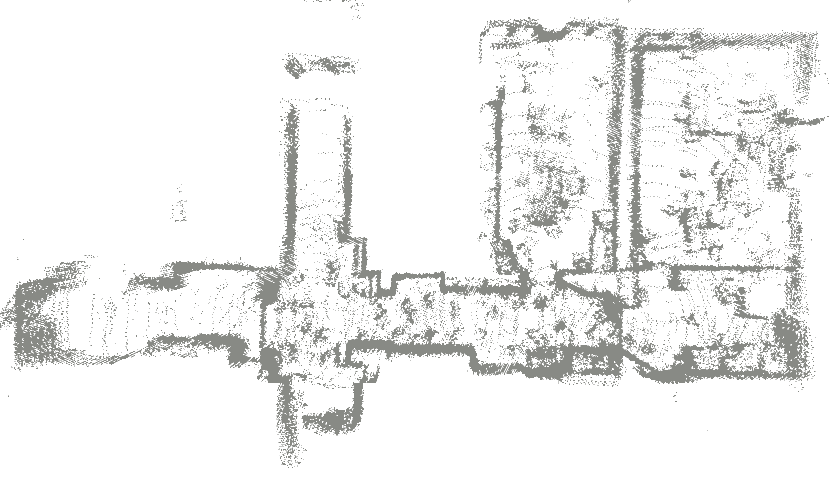}
         \caption{\textit{S-Graphs+} on \textit{Seq-06}}
         \label{fig_eval_sgraphs_s6}
     \end{subfigure}
     \hfill
     \begin{subfigure}[t]{0.23\textwidth}
         \centering
         \includegraphics[width=\textwidth]{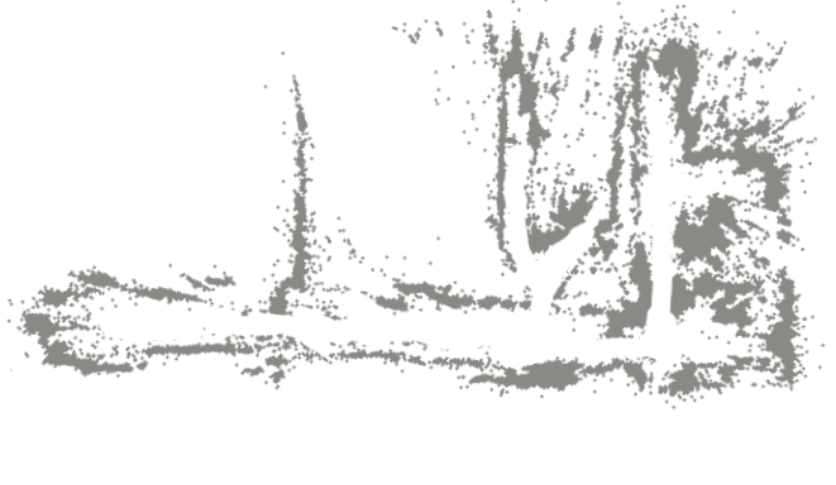}
         \caption{UcoSLAM \cite{munoz2020ucoslam} on \textit{Seq-06}}
         \label{fig_eval_uco_s6}
     \end{subfigure}
     \hfill
     \begin{subfigure}[t]{0.23\textwidth}
         \centering
         \includegraphics[width=\textwidth]{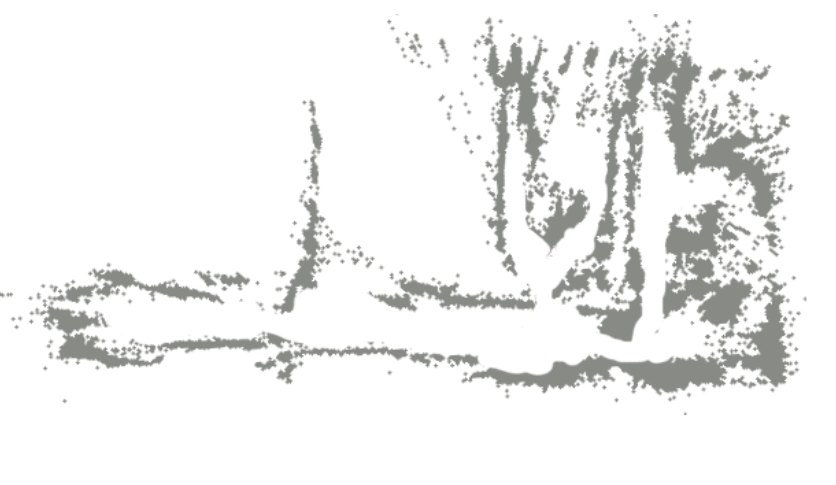}
         \caption{Proposed method on \textit{Seq-06}}
         \label{fig_eval_proposed_s6}
     \end{subfigure}
     \hfill
     \begin{subfigure}[t]{0.23\textwidth}
         \centering
         \includegraphics[width=\textwidth]{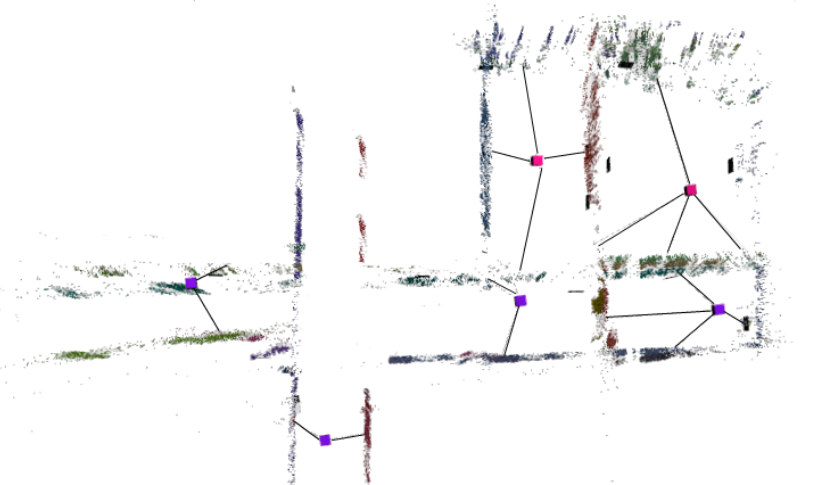}
         \caption{Proposed method on \textit{Seq-06} showing only wall 3D points}
         \label{fig_eval_semantic_s6}
     \end{subfigure}
     \caption{Reconstructed maps using the proposed method, the baseline, and the ground truth. It can be seen that the proposed approach is able to generate more precise maps similar to the ground truth in detailed and with high-level semantic entities.}
     \label{fig_evals}
\end{figure*}

The above-mentioned impact is more obvious in cases such as \textit{Seq-04} where the robot starts in a corridor, enters from one of the two doors of a room and exits from the other door to continue in the same corridor, and as a result, no loop closure using keypoints and markers is performed.
While our method, which creates wall and room constraints, correctly identifies the corridor and its walls to provide better results.
This improvement can be clearly seen in Fig.~\ref{fig_evals}, where our method is able to reconstruct a more accurate map of the environment when compared with the baseline.
Moreover, it is able to extract meaningful semantic and topological information from the environment.

\begin{table}[t]
    \small
    \centering
    \caption{Evaluation results of the proposed method on the collected dataset using \acf{RMSE} error in \textit{meters} and \acf{STD}. The best results are boldfaced.}
    \begin{tabular}{l | cc | cc}
        \toprule
            & \multicolumn{4}{c}{\textbf{Method}} \\
            \cmidrule{2-5}
            & \multicolumn{2}{c|}{Proposed} & \multicolumn{2}{c}{UcoSLAM \cite{munoz2020ucoslam}} \\
            \midrule
        \multicolumn{1}{l|}{\textbf{Sequence}}         & RMSE           & \multicolumn{1}{c|}{STD}            & RMSE             & STD              \\
        \midrule
        \multicolumn{1}{l|}{\textit{Seq-01}} & \textbf{8.038} & \multicolumn{1}{c|}{3.058}          & 8.130            & \textbf{3.035}   \\
        \multicolumn{1}{l|}{\textit{Seq-02}} & \textbf{6.883} & \multicolumn{1}{c|}{\textbf{3.598}} & 6.930            & 3.633            \\
        \multicolumn{1}{l|}{\textit{Seq-03}} & \textbf{2.266} & \multicolumn{1}{c|}{\textbf{0.894}} & 2.687            & 1.335            \\
        \multicolumn{1}{l|}{\textit{Seq-04}} & \textbf{3.787} & \multicolumn{1}{c|}{\textbf{1.848}} & 5.822            & 2.726            \\
        \multicolumn{1}{l|}{\textit{Seq-05}} & 1.255          & \multicolumn{1}{c|}{0.763}          & \textbf{1.238}   & \textbf{0.751}   \\
        \multicolumn{1}{l|}{\textit{Seq-06}} & \textbf{2.676} & \multicolumn{1}{c|}{\textbf{1.524}} & 2.720            & 1.536            \\
        \bottomrule
    \end{tabular}
    \label{tbl_evaluation}
\end{table}

\textbf{Limitations.} The proposed method, when testing in an environment where marker-based loop closure is triggered several times, can show similar results to baseline UcoSLAM.
\textit{Seq-06} in Fig.~\ref{fig_evals} shows an example of the final map showcasing similar map quality results generated by the proposed method and UcoSLAM for an environment where the robot revisits the places (thus the markers) several times.
Even with the given limitation for such scenarios, our approach is able to extract more meaningful information from the environment while maintaining similar accuracy with respect to the baseline.       

\section{Conclusions}
\label{sec_conclusions}

This paper presented a \acl{VSLAM} approach that employs a monocular camera as the sensor and the data encoded within fiducial markers placed in the surroundings for accurate pose estimation and semantic segmentation.
The proposed method can detect three practical semantic entities in the environment, including walls, corridors, and rooms, and provide a hierarchical graph with high-level semantic data.
Additionally, it can reconstruct a more accurate map of the environment by adding more topological constraints and employing the relations among mentioned semantic entities found in the scene.
Evaluations performed on a dataset collected by a legged robot in real-world conditions demonstrated an accuracy improvement compared to the baseline and satisfactory results compared to a \ac{LiDAR}-based state-of-the-art techniques.

The approach proposed in this paper is part of a project which utilizes the advantages of fiducial markers for various tasks while keeping these markers invisible to human eyes but recognizable to robots.
As for future works, the authors plan to encode the environment with the mentioned invisible fiducial markers introduced in \cite{agha2022unclonable} and equip the robot with a proper sensor setup to interpret them, making the improved version of the proposed approach able to accomplish \acl{SLAM} tasks with the new type of markers.
Additionally, improving the performance of the proposed system to work online and in real-time is another target for future work.



\bibliographystyle{IEEEtran}
\bibliography{root}

\end{document}